\newcommand{\CLASSINPUTtoptextmargin}{0.75in}
\newcommand{\CLASSINPUTbottomtextmargin}{1.02in}

\documentclass[conference]{IEEEtran}

\IEEEoverridecommandlockouts

\usepackage{cite}
\usepackage{amsmath,amssymb,amsfonts}
\usepackage{algorithmic}
\usepackage{graphicx}
\usepackage{textcomp}
\usepackage{xcolor}
\usepackage{booktabs}
\usepackage{graphicx}
\usepackage{multirow}
\usepackage{algorithm}

\begin{document}

\title{DentAgent: Evidence-Centric Multi-Agent Coordination for Multimodal Dental Reasoning}

\author{
Zijie Meng\textsuperscript{1,$\star$}\thanks{$^\star$Equal contribution.}, 
Xiwei Dai\textsuperscript{1,$\star$}, 
Yixuan Tang\textsuperscript{1},
Jin Hao\textsuperscript{2},
Yang Feng\textsuperscript{3},
Fudong Zhu\textsuperscript{1},\\
Xiaoqiang Liu\textsuperscript{4},
Shaosheng Cao\textsuperscript{5}, 
Zuozhu Liu\textsuperscript{1,\dag}\thanks{$^\dagger$Corresponding author.} \\
\textsuperscript{1}\textit{Zhejiang University},
\textsuperscript{2}\textit{Shanghai Jiao Tong University},
\textsuperscript{3}\textit{Angelalign Technology Inc.}, \\
\textsuperscript{4}\textit{Peking University}, 
\textsuperscript{5}\textit{Tsinghua University} \\
}


\maketitle

\begin{abstract}
Oral diseases affect billions of people worldwide, underscoring a pressing need for accurate and reliable dental assessment that integrates heterogeneous evidence from domain knowledge, radiographs, intraoral photographs, and 3D dental data. Most existing dental AI systems remain modality- or task-specific. Although recent vision-language models support flexible dental question answering, directly generated response leaves evidence implicit and untraceable. To address these limitations, we introduce DentAgent, an evidence-centric multi-agent framework, in which the Orchestrator coordinate five specialized agents spanning various modalities. Each specialist utilizes domain tools to convert observations into structured evidence records. The Evidence Blackboard manages these records as a shared evidence state, tracking coverage, gaps, and conflicts before response generation. This standardized evidence representation integrates isolated dental capabilities into a unified agentic workflow. Across four benchmarks, DentAgent demonstrates leading performance, even surpassing the senior specialists by 17.3 percentage points on multi-label diagnosis, which supports its value for broadly applicable and traceable multimodal dental reasoning, and highlights its potential as a technical foundation for population oral health assessment and management.
\end{abstract}


\section{Introduction}
Oral diseases affect billions of people worldwide and impose substantial health and socioeconomic burdens \cite{who2025oralhealth,peres2019oral,watt2019ending}. Effective prevention and treatment depend in part on accurate and reliable dental assessment, which often draws on heterogeneous evidence, including domain knowledge, radiographs, intraoral photographs, and 3D dental data. However, these sources differ substantially in representation, anatomical granularity, and evidential role, requiring different forms of perception, measurement, and reasoning. This heterogeneity creates a central challenge for dental AI: a broadly applicable system must coordinate modality-specific capabilities while making explicit which observations support each conclusion.

Recent dental AI systems have made substantial progress in addressing individual parts of this problem, ranging from task-specific perception and measurement models to domain-oriented foundation and generative models \cite{schwendicke2020ai,abusalim2022analysis,huang2023chatgpt,meng2025dentvlm,huang2025dentvfm,zhu2026dentfound}. However, most existing systems remain organized around individual modalities or narrowly defined tasks. Specialized models provide useful local capabilities but typically operate as isolated pipelines, whereas generative models often map their inputs directly to answers or reports. In both cases, the evidence supporting the model output generally remains implicit, making it difficult to trace a prediction or response to specific visual findings, retrieved knowledge, or quantitative measurements.

Tool-augmented and agentic reasoning provides a natural alternative to direct generation by allowing language models to invoke specialized tools and reason over their outputs as intermediate observations \cite{yao2023react,schick2023toolformer,shen2023hugginggpt}. Recent dental agents have applied this paradigm to selected imaging modalities and dental workflows \cite{hao2026oralagent,yu2026opgagent,hao2026orthoagent}. But these systems remain largely modality- or task-specific. More importantly, tool use alone does not provide a common mechanism for coordinating heterogeneous capabilities, standardizing their outputs, linking them to their sources, or explicitly representing missing and conflicting evidence. A broader dental reasoning framework therefore requires not only specialized tools but also a structured evidence representation designed to connect modality-specific observations to final answer generation.

To address these limitations, we propose DentAgent, an evidence-centric multi-agent framework that coordinates heterogeneous dental capabilities within a unified workflow. DentAgent primarily consists of an Orchestrator, five Modality-specific Specialists, and an Evidence Blackboard. The Orchestrator identifies response-critical evidence awaiting acquisition and dynamically activates the relevant specialist agent. Each agent follows a bounded loop, iteratively selecting tools, examining observations, and converting them into structured evidence records. The Evidence Blackboard normalizes these evidence, tracks their coverage, links them across agents, and deals unresolved gaps or conflicts. Based on the evolving evidence state, the Orchestrator may request further analysis before committing the final answer. 

We evaluate DentAgent on four benchmarks spanning bilingual dental knowledge \cite{zhu2025dentalbench}, panoramic radiograph question answering (QA) \cite{hao2025mmoral}, 2D dental diagnosis \cite{meng2025dentvlm}, and 3D intraoral scan (IOS) reasoning \cite{xiong2026iosvlm}, where DentAgent obtains the leading scores in various scenarios, especially exceeding the senior specialists by 17.3 percentage points on multi-label diagnosis. Together, these results support the applicability and traceability of DentAgent as a unified framework for dental reasoning, while highlighting its potential as a technical foundation for population oral health assessment and management.

Our contributions are summarized as follows:

\begin{itemize}
    \item We introduce DentAgent, a multi-agent framework that coordinates specialist capabilities for textual knowledge, radiographs, clinical photography, and 3D dental data under a unified orchestration protocol.

    \item We develop an evidence-centric coordination mechanism that decouples evidence acquisition from answer generation, enabling modality-specific specialists to contribute source-attributed findings to a shared Evidence Blackboard for target-driven reasoning and iterative evidence-sufficiency assessment.

    \item We conduct evaluations on heterogeneous dental benchmarks, demonstrating the applicability of DentAgent across diverse modalities and tasks, with clear improvements over strong baselines.
\end{itemize}

\section{Related Works}

\subsection{Task-Specific Dental Perception Models}
Early dental AI research primarily addressed well-defined perception tasks within individual imaging modalities. Deep learning has been applied to caries detection on periapical and bitewing radiographs \cite{lee2018caries,cantu2020caries}, periodontal bone-loss detection and measurement \cite{krois2019periodontal,lee2021alveolar}, tooth detection and numbering in panoramic radiographs \cite{tuzoff2019tooth}, and instance-level tooth segmentation \cite{jader2018deep}. In orthodontics, public cephalometric challenges and automated landmarking systems established important benchmarks for landmark localization and measurement-based analysis \cite{wang2016benchmark,lindner2016ceph}. For 3D dental data, ToothNet, CBCT segmentation systems, MeshSNet, and TSegNet further demonstrate the importance of explicit geometric and spatial modeling for tooth- and surface-level understanding \cite{cui2019toothnet,cui2022cbct,lian2019meshsnet,lian2020mesh,cui2021tsegnet,shi2024leta}. These studies provide strong perceptual capabilities for specific modalities and clinical tasks, but they generally operate in isolation and do not address how heterogeneous findings should be aligned, reconciled, and integrated across data sources.

\subsection{Dental Foundation and Generative Models}
Recent studies have expanded dental AI beyond narrowly defined prediction tasks toward domain-specialized foundation and generative models. OralGPT introduces large-scale instruction data and evaluation protocols for panoramic X-ray understanding, while DentalBench provides a bilingual benchmark for evaluating dental knowledge in LLMs \cite{hao2025mmoral,zhu2025dentalbench}. DentVLM extends vision-language modeling across multiple 2D oral imaging modalities and diagnostic tasks, whereas DentVFM investigates scalable vision foundation models for oral and maxillofacial radiology \cite{meng2025dentvlm,huang2025dentvfm}. DentFound further applies instance-guided vision-language modeling to panoramic radiograph interpretation and report generation \cite{zhu2026dentfound}. Beyond 2D imaging, IOSVLM models the native geometry of 3D intraoral scans, CBCTRepD addresses bilingual report generation from CBCT data, and dental NLP studies explore evidence extraction and documentation from clinical notes \cite{xiong2026iosvlm,wu2026cbctrepd,pethani2023nlp,buttner2024nlp}. Collectively, these works demonstrate the value of dental-specific data and modality-aware representation learning \cite{hao2026oralgpt,cai2026dentalgpt,zhang2025archmap}; however, they remain largely centered on the capabilities of individual models and provide limited support for explicitly coordinating and managing evidence produced by multiple specialized components.

\subsection{Agentic Dental AI}
Building on advances in dental foundation and generative models, recent work has begun to shift from enhancing individual model capabilities toward coordinating multiple sources of specialized expertise. OralAgent integrates visual tools and dental knowledge retrieval for interactive dental image analysis \cite{hao2026oralagent}, while OPGAgent orchestrates specialized perception modules, hierarchical evidence collection, and consensus-based reporting for panoramic radiograph interpretation \cite{yu2026opgagent}. OralGPT-Plus further investigates iterative and symmetry-aware tool use for panoramic X-ray analysis through reinforcement learning \cite{fan2026oralgptplus}, and orthodontic agent systems demonstrate the potential of role-specialized collaboration for diagnosis and treatment planning \cite{hao2026orthoagent}. These studies establish the value of tool-augmented and collaborative reasoning in dentistry, but they are generally designed around a particular imaging modality or clinical workflow. DentAgent extends this direction to a broader multimodal setting, in which specialists for clinical text, radiographs, clinical photography, and 3D dental data contribute complementary observations to a shared evidence blackboard, enabling structured evidence coordination and traceable response generation.

\begin{figure*}[t]
    \centering
    \includegraphics[width=\textwidth]{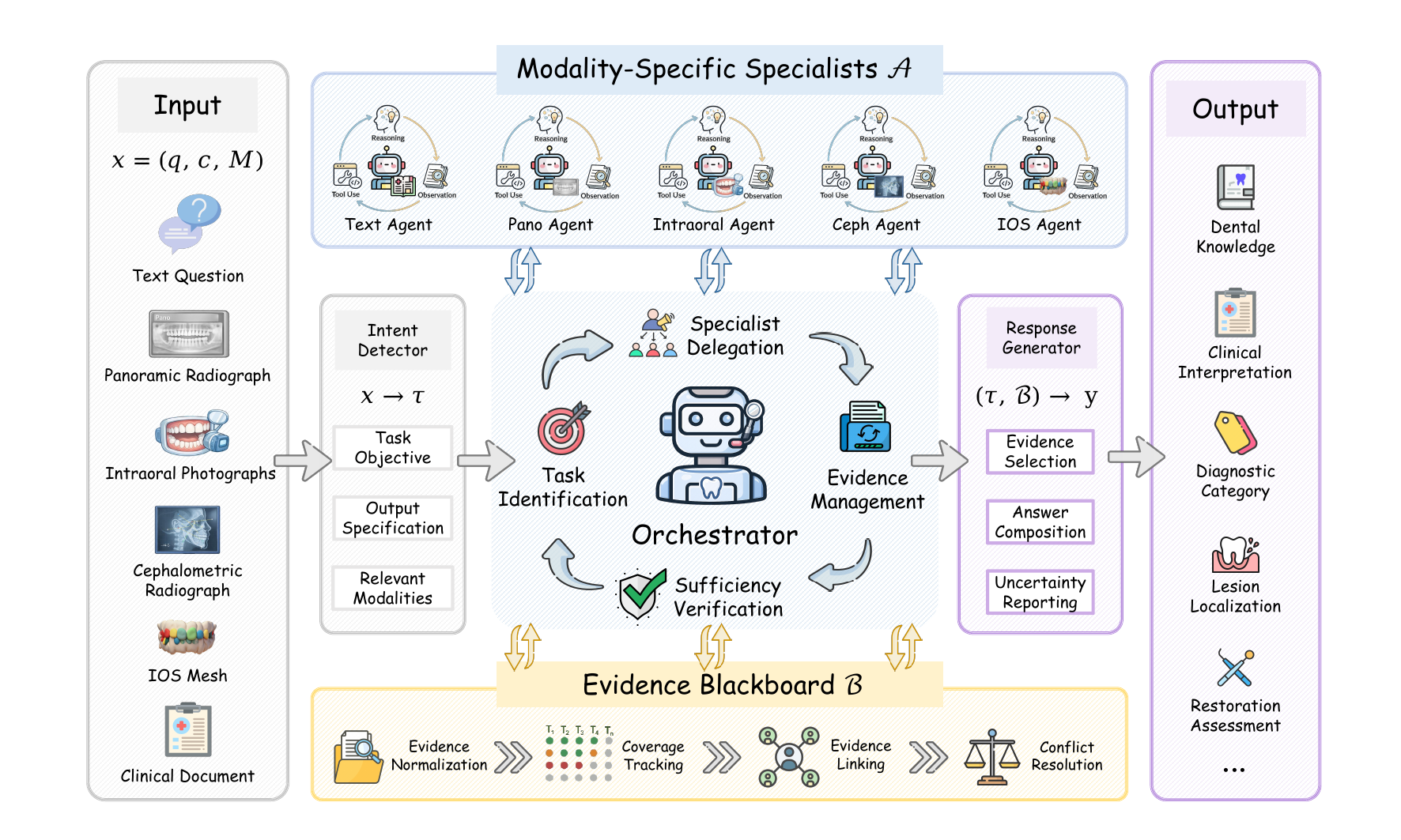}
    \caption{Overview of DentAgent, an evidence-centric hierarchical multi-agent framework for multimodal dental reasoning. Given a dental case \(x=(q,c,\mathcal{M})\), where \(q\), \(c\) and \(\mathcal{M}\) denote the query, clinical context and multimodal inputs respectively, the Intent Detector interprets the requested tasks, after which the Orchestrator iteratively identifies task-relevant evidence targets, delegates them to appropriate Modality-Specific Specialists, manages the updated evidence state, and verifies its sufficiency. Each activated specialist performs a bounded reasoning, tool-use and observation cycle. The resulting observations are incorporated into the Evidence Blackboard by evidence normalization, coverage tracking, evidence linking, and conflicts resolution. Finally, the Response Generator maps the terminal evidence state to the requested output, including dental knowledge, diagnostic category, lesion localization, or restoration assessment.}
    \label{fig:dentagent}
\end{figure*}

\section{Methods}
\label{sec:methods}

\subsection{Framework Overview}
\label{sec:framework_overview}

DentAgent is an evidence-centric hierarchical multi-agent framework for multimodal dental reasoning. Given a dental case \(x=(q,c,\mathcal{M})\), where \(q\) denotes the task query, \(c\) denotes optional clinical context, and \(\mathcal{M}\) contains the available multimodal inputs, DentAgent produces a task-appropriate response \(y\) grounded in the evidence available in the case. The framework supports text questions, clinical documents, panoramic and cephalometric radiographs, intraoral photographs, and IOS meshes. Its output space covers dental knowledge answering, diagnostic category prediction, lesion localization, and restoration assessment.

DentAgent separates task interpretation, specialist execution, evidence organization, and response generation. The \emph{Intent Detector} first maps the input case to a task representation \(\tau\), which specifies the concrete task requested and the expected form and granularity of the response. The \emph{Orchestrator} then follows the loop from task identification, specialists delegation, evidence management to sufficiency verification, as illustrated in Fig.~\ref{fig:dentagent}. Rather than directly producing a diagnosis, it identifies the evidence targets required by the task, assigns them to an appropriate subset of \emph{Modality-Specific Specialists}, and repeatedly evaluates the resulting evidence state.

Each specialist operates within a bounded modality-specific reasoning, tool-use, and observation cycle. Each specialist activation is limited to at most \(K_{\max}\) tool calls, thereby preventing unbounded or non-terminating tool-use loops. The resulting observation packets are incorporated into the \emph{Evidence Blackboard} through four successive operations: \emph{Evidence Normaization}, \emph{Coverage Tracking}, \emph{Evidence Linking}, and \emph{Conflict Resolution}. The updated blackboard is returned to the Orchestrator, enabling subsequent rounds to focus on unresolved or contradictory evidence rather than repeating completed analyses. Execution terminates when verification succeeds, no additional evidence target is identified, or the maximum number of global rounds \(T_{\max}\) is reached. The \emph{Response Generator} then maps the terminal blackboard state to the requested output. Algorithm~1 summarizes the complete inference procedure.


\begin{algorithm}[t]
\caption{Evidence-Centric Inference with DentAgent}
\label{alg:dentagent}
\begin{algorithmic}[t]
\REQUIRE Dental case \(x=(q,c,\mathcal{M})\); specialist set \(\mathcal{A}=\{\mathcal{A}_k\}_{k=1}^{N}\); specialist tool inventories \(\mathcal{T}=\{\mathcal{T}_k\}_{k=1}^{N}\); maximum global rounds \(T_{\max}\); maximum tools per activation \(K_{\max}=3\)
\ENSURE Final response \(y\)
\STATE \(\tau \gets \mathrm{IntentDetector}(x)\)
\STATE \(\mathcal{B}^0 \gets \mathrm{InitializeBlackboard}()\)
\FOR{\(t=1\) to \(T_{\max}\)}
    \STATE \(\mathcal{R}^{(t)} \gets \mathrm{IdentifyTargets}(\tau,\mathcal{B}^{(t-1)})\)
    \IF{\(\mathcal{R}^{(t)}=\varnothing\)}
        \STATE \(\mathcal{B}^{(t)} \gets \mathcal{B}^{(t-1)}\)
        \STATE \textbf{break}
    \ENDIF
    \STATE \(\mathcal{D}^{(t)} \gets \mathrm{DelegateSpecialists}(\mathcal{R}^{(t)},\mathcal{B}^{(t-1)},\mathcal{A})\)
    \STATE \(\mathcal{O} \gets \varnothing\)
    \FOR{each assignment \((\mathcal{A}_k,\mathcal{R}^{(t)}_k)\in\mathcal{D}^{(t)}\)}
        \STATE \(\widehat{\mathcal{T}}_k \gets \mathrm{SelectTools}(\mathcal{A}_k,\mathcal{R}^{(t)}_k,\mathcal{B}^{(t-1)},\mathcal{T}_k,K_{\max})\)
        \STATE \(\mathcal{O}_k \gets \mathrm{ExecuteSpecialist}(\mathcal{A}_k,\mathcal{R}^{(t)}_k,\widehat{\mathcal{T}}_k,x,\mathcal{B}^{(t-1)})\)
        \STATE \(\mathcal{O} \gets \mathcal{O}\cup\{\mathcal{O}_k\}\)
    \ENDFOR
    \STATE \(\mathcal{B}^{(t)} \gets \mathrm{NormalizeEvidence}(\mathcal{B}^{(t-1)},\mathcal{O})\)
    \STATE \(\mathcal{B}^{(t)} \gets \mathrm{TrackCoverage}(\mathcal{B}^{(t)},\mathcal{R}^{(t)})\)
    \STATE \(\mathcal{B}^{(t)} \gets \mathrm{LinkEvidence}(\mathcal{B}^{(t)})\)
    \STATE \(\mathcal{B}^{(t)} \gets \mathrm{ResolveConflicts}(\mathcal{B}^{(t)})\)
    \IF{\(\mathrm{Verify}(\tau,\mathcal{B}^{(t)})\)}
        \STATE \textbf{break}
    \ENDIF
\ENDFOR
\STATE \(y \gets \mathrm{GenerateResponse}(\tau,\mathcal{B}^{(t)})\)
\STATE \textbf{return} \(y\)
\end{algorithmic}
\end{algorithm}

\subsection{Global Orchestration}
\label{sec:global_orchestration}
The evidence required for multimodal dental reasoning is inherently task-dependent. Dental knowledge questions rely primarily on textual information, whereas diagnostic classification, lesion localization, and restoration assessment require different combinations of visual and geometric evidence at varying levels of granularity. To accommodate these task-specific requirements, DentAgent separates task interpretation from evidence acquisition: the Intent Detector first determines what must be answered, and the Orchestrator subsequently determines which evidence must be acquired to support that answer.

At the start of inference, the Intent Detector maps the input case \(x\) to a task specification:
\begin{equation}
\tau = \mathrm{IntentDetector}(x),
\end{equation}
where the specification \(\tau\) defines the task objective, required output format and granularity, and the input modalities relevant to the task. It contains neither a diagnostic prediction nor a candidate response; instead, it remains fixed throughout inference and guides subsequent rounds in identifying evidence targets, acquiring the required evidence, and verifying its sufficiency for the requested output.

The Orchestrator then follows the \emph{Identification--Delegation--Management--Verification} loop shown in Fig.~\ref{fig:dentagent}. At each orchestration round \(t\), the \emph{Task Identification} stage derives the evidence targets that remain unresolved:
\begin{equation}
\mathcal{R}^{(t)} =
\mathrm{IdentifyTargets}
\left(
\tau,
\mathcal{B}^{(t-1)}
\right),
\end{equation}
where \(\mathcal{B}\) denotes the Evidence Blackboard, which manages the shared evidence state across orchestration rounds. Each target $\mathcal{R}$ denotes a task-specific evidence requirement, such as the clinical finding, anatomical location, quantitative measurement, structural relationship, or knowledge claim. The evidence required for a target is determined by the requested output. For example, evidence sufficient to support a coarse diagnostic category may not provide the spatial or quantitative detail needed for precise localization or measurement. After each blackboard update, the Orchestrator reassesses all targets, removes those that have been adequately resolved, and retains those that remain incomplete, unsupported, or contested.

During the \emph{Specialist Delegation} stage, the Orchestrator assigns the current targets to an appropriate subset of Specialist Sub-Agents:
\begin{equation}
\mathcal{D}^{(t)}
=
\mathrm{DelegateSpecialists}
\left(
\mathcal{R}^{(t)},
\mathcal{B}^{(t-1)},
\mathbf{A}
\right).
\end{equation}
where each assignment \((\mathcal{A}_k,\mathcal{R}_k)\in\mathcal{D}^{(t)}\) associates specialist \(\mathcal{A}_k\) with a target subset \(\mathcal{R}_k\subseteq\mathcal{R}^{(t)}\). Assignments are determined by the required evidence type, available case inputs, and specialist capability boundaries. The Orchestrator may activate a single specialist for a modality-specific target or multiple complementary specialists when evidence from several modalities is required. 

After the selected specialists return their current-round outputs, the Orchestrator enters the \emph{Evidence Management} stage, where the Evidence Blackboard integrates and manages the newly acquired evidence. It then performs $\mathrm{Verify}\left(\tau,\mathcal{B}^{(t)}\right)$ to determine whether the current evidence state is sufficient for the requested output. If verification succeeds, evidence acquisition terminates. Otherwise, the updated blackboard conditions the next \emph{Identify Task} stage, which derives a revised target set \(\mathcal{R}^{(t+1)}\), directing subsequent execution toward the remaining evidence gaps and conflicts. Thus, specialist routing adapts to the accumulated evidence, while the task specification \(\tau\) remains fixed throughout inference.

\subsection{Specialist Sub-Agent Execution}
\label{sec:specialist_execution}

DentAgent includes five Specialist Sub-Agents, each designed for a specific modality of clinical data. The \emph{Text Agent} handles clinical text and dental knowledge; the \emph{Pano Agent} analyzes panoramic radiographs; the \emph{Intraoral Agent} examines intraoral photographs; the \emph{Ceph Agent} analyzes cephalometric radiographs; and the \emph{IOS Agent} analyzes IOS based dental and occlusal structures. These specialists have complementary roles and are activated as needed, without a fixed execution order.

Each specialist \(\mathcal{A}_k\) is associated with a corresponding tool inventory \(\mathcal{T}_k\). Given assigned targets \(\mathcal{R}_k\), it selects at most \(K_{\max}\) relevant tools:
\begin{equation}
\begin{aligned}
\widehat{\mathcal{T}}_k
&=
\mathrm{SelectTools}
\left(
\mathcal{A}_k,\mathcal{R}_k,\mathcal{B},
\mathcal{T}_k,K_{\max}
\right),\\
\widehat{\mathcal{T}}_k
&\subseteq \mathcal{T}_k,
\qquad
\left|\widehat{\mathcal{T}}_k\right|
\leq K_{\max},
\end{aligned}
\end{equation}
where tool selection depends on the assigned targets, available inputs, current blackboard state, and the relevance of each tool.

Then, the specialist executes the selected tools and returns their observations:
\begin{equation}
\mathcal{O}_k
=
\mathrm{ExecuteSpecialist}
\left(
\mathcal{A}_k,
\mathcal{R}_k,
\widehat{\mathcal{T}}_k,
x,
\mathcal{B}
\right).
\end{equation}
The output \(\mathcal{O}_k\) contains findings or measurements for the assigned targets, together with their anatomical location, source, quality, and conditions of validity.

\subsection{Evidence Blackboard}
\label{sec:evidence_blackboard}

The Evidence Blackboard \(\mathcal{B}\) stores and organizes all evidence collected during inference. It serves as the shared state used by the Specialist Sub-Agents, the Orchestrator, and the Response Generator. Evidence from different specialists and reasoning rounds is aligned to the same target while explicitly record their source and applicability conditions.

\paragraph{Evidence Normalization.}
\(\mathrm{NormalizeEvidence}\) converts each specialist output into a standardized evidence record. The process aligns clinical terminology, anatomical labels, and levels of detail, allowing evidence from different specialists to be compared directly. Each record specifies the target, finding or measurement, anatomical location, originating specialist, source tool, quality, and limitations. The resulting representation supports subsequent coverage assessment and evidence linking from different sub-agents.

\paragraph{Coverage Tracking.}
\(\mathrm{TrackCoverage}\) assesses how well the current evidence addresses each target. It assigns or updates one of four states, including covered, partially covered, conflicting, or uncovered. The required evidence depends on the requested output. For example, coarse findings may be sufficient to confirm that an abnormality is present, but insufficient to determine its exact location, severity, or size.

\paragraph{Evidence Linking.}
\(\mathrm{LinkEvidence}\) links evidence records associated with the same target and classifies their relationships as supporting, complementary, redundant, or conflicting. Supporting records provide independent evidence for the same conclusion; complementary records describe different aspects of the target; redundant records repeat an existing observation; and conflicting records report incompatible findings. Linked records are retained separately rather than merged into a single record. This representation preserves source-specific information and indicates whether a conclusion is supported by multiple independent observations or by a single source.

\paragraph{Conflict Resolution.}
\(\mathrm{ResolveConflicts}\) evaluates incompatible evidence records according to target relevance, anatomical correspondence, source applicability, observation quality, provenance, and applicability conditions. Resolution does not rely on majority voting. A record may receive greater evidential weight only when it evaluates the target more directly and reliably. All source records remain retained for traceability. When the available evidence does not justify a reliable preference, the disagreement remains explicit. This operation also updates the coverage states of the affected targets before the blackboard is returned to the Orchestrator.

\subsection{Response Generation}
\label{sec:response_generation}
After the orchestration loop terminates, the Response Generator produces the final output from the task specification \(\tau\) and the final blackboard state \(\mathcal{B}\):
\begin{equation}
y=\mathrm{GenerateResponse}(\tau,\mathcal{B}).
\end{equation}
The Response Generator does not call additional tools or modify the blackboard. It selects the evidence relevant to the requested targets and presents the result according to \(\tau\). For closed-set tasks, the output is restricted to the predefined label set. For structured and open-ended tasks, the response is generated from the evidence records stored in the blackboard. If \(\tau\) defines uncertainty as a valid output, the Response Generator reports unresolved conflicts or insufficient evidence instead of forcing a definitive answer.

\section{Experiments}
\subsection{Implementation}
DentAgent is implemented in LangGraph, with a shared blackboard that manages accumulated evidence and execution status across orchestration rounds. Unless specified, Qwen3.5-9B \cite{qwen3.5}, served through vLLM \cite{vllm}, is used for both orchestration and final response generation with tailored system prompts. Reasoning mode is enabled during evidence acquisition and disabled during final response generation. The framework includes \(N=5\) modality-specific specialists and $33$ tools. Each specialist is registered through a YAML capability card that specifies its tool inventory, supported modality, target coverage, input--output schema, anatomical scope, and applicability constraints. During each orchestration round, compatible specialists can be executed in parallel, and each specialist invocation may select at most \(K_{\max}=3\) tools. We set the global execution budget to \(T_{\max}=15\) rounds and terminates earlier when verification succeeds or no feasible specialist remains.

\subsection{Benchmarks}
We evaluate DentAgent on four benchmarks spanning bilingual dental textual knowledge, 2D clinical photographs, panoramic radiograph interpretation, and IOS based 3D reasoning. These benchmarks cover closed-set classification, structured prediction, and open-ended clinical question answering. Unless specified, we follow the official evaluation protocols and score only the final response. 

\paragraph{DentalBench \cite{zhu2025dentalbench}} It comprises 7,332 English and Chinese questions across multiple dental specialties, including 5,408 close-ended and 1,924 open-ended questions. Following the official partition, we report accuracy and BERTScore \cite{zhang2020bertscore} for close-ended and open-ended questions, respectively. 

\paragraph{MMOralBench~\cite{hao2025mmoral}} It evaluates clinical reasoning based on panoramic radiographs. We use the open-ended subset of 578 questions, with each response evaluated against the ground-truth answer by GPT-5-mini~\cite{gpt5} and assigned a score in $[0,1]$.

\paragraph{DentVLM Reader-study Benchmark \cite{meng2025dentvlm}} It contains 3,105 diagnostic questions grounded in panoramic radiographs, intraoral photographs, and lateral cephalometric radiographs. It evaluates 36 types of clinical findings, where multi-class tasks are evaluated by exact-match accuracy, while multi-label tasks are evaluated by hit rate.

\paragraph{IOSVQA~\cite{xiong2026iosvlm}} It evaluates 3D dental reasoning based on IOS, which includes 1,929 questions covering eight diagnostic tasks. Following the original protocol, the performance is measured by accuracy.

\subsection{Main Results}
\begin{table}[t]
    \centering
    \caption{Comparison on DentalBench across textual closed- and open-ended question answering.}
    \label{tab:dentalbench_results}
    \resizebox{.95\linewidth}{!}{
    \begin{tabular}{lcc}
        \toprule
        \multirow{2}{*}{\textbf{Model}} & \textbf{close-ended} & \textbf{Open-ended} \\
        {} & \textit{Accuracy} (\%) $\uparrow$ & \textit{BERTScore} (\%) $\uparrow$ \\
        \midrule
        \multicolumn{3}{@{}l}{\textit{General-purpose LLMs}} \\
        DeepSeek-R1
        & \textbf{68.44}
        & 20.71 \\
        DeepSeek-V3
        & \underline{67.03}
        & 25.15 \\
        GPT-4o
        & 66.05
        & 27.80 \\
        GPT-4o-mini
        & 53.16
        & 28.32 \\
        Qwen2.5-32B
        & 64.49
        & 22.44 \\
        Qwen2.5-14B
        & 58.84
        & 22.38 \\
        Qwen2.5-7B
        & 54.23
        & 22.26 \\
        
        \midrule
        \multicolumn{3}{@{}l}{\textit{Dental-adapted LLMs}} \\
        Qwen2.5-3B + SFT
        & 50.35
        & 27.81 \\
        Qwen2.5-3B + RAG
        & 50.26
        & 30.54 \\
        Qwen2.5-3B + SFT + RAG
        & 55.27
        & \underline{30.57} \\
        
        \midrule
        \textbf{DentAgent (ours)}
        & 66.32
        & \textbf{33.31} \\
        \bottomrule
    \end{tabular}
    }
\end{table}

\paragraph{DentalBench}
As shown in Table~\ref{tab:dentalbench_results}, DentAgent achieves the best open-ended result, with a BERTScore of 33.31\%, outperforming the strongest dental-adapted baseline by 2.74\%. On close-ended questions, DentAgent attains an accuracy of 66.32\%, trailing DeepSeek-R1 by 2.12\%, while outperforming GPT-4o and all dental-adapted baselines. In contrast to predefined-label selection in close-ended questions, open-ended tasks require models to generate comprehensive responses that explain specialized dental concepts, analyze clinical cases, or address patient needs. Consequently, open-ended tasks place greater demands on dental knowledge and evidence retrieval and integration, highlighting DentAgent's ability to coordinate its text agent and evidence management module to produce well-supported answers.

\begin{table}[t]
    \centering
    \caption{Comparison with various baselines on MMOralBench. DentAgent is implemented based on GPT-5.4-mini. Responses are scored by GPT-5-mini against reference answers.}
    \label{tab:mmoral_overall}
    \resizebox{.75\linewidth}{!}{%
    \begin{tabular}{lc}
        \toprule
        \textbf{Model} & \textbf{Model-Judge Score (\%) $\uparrow$} \\
        \midrule
        \multicolumn{2}{@{}l}{\textit{General-purpose MLLMs}} \\
        GPT-5                          & 42.42 \\
        GPT-4V                         & 39.38 \\
        Gemini-2.5-Flash               & 27.84 \\
        Qwen-Max-VL                    & 5.29 \\
        DeepSeek-VL-7B-Chat            & 15.95 \\
        GLM-4V-9B-Thinking             & 19.74 \\
        Qwen2.5-VL-72B                 & 15.38 \\
        \midrule
        
        \multicolumn{2}{@{}l}{\textit{Medical-specific MLLMs}} \\
        LLaVA-Med                      & 4.76 \\
        HealthGPT-XL32                 & 27.80 \\
        MedVLM-R1                      & 24.70 \\
        MedDr                          & 26.20 \\
        OralGPT-Omni                   & 45.31 \\
        \midrule
        
        \multicolumn{2}{@{}l}{\textit{Medical agents}} \\
        MedRAX                         & 36.73 \\
        MedAgents                      & 34.71 \\
        MMedAgent                      & 15.19 \\
        MDAgents                       & 40.50 \\
        OralAgent                      & \underline{61.00} \\
        \midrule
        \textbf{DentAgent (ours)}      & \textbf{61.44} \\
        \bottomrule
    \end{tabular}
    }
\end{table}

\paragraph{MMOralBench}
As shown in Table~\ref{tab:mmoral_overall}, for question answering based on panoramic radiographs, DentAgent outperforms OralGPT-Omni, the strongest dental-specific MLLM among the evaluated baselines, by 16.13\%, while also surpassing all evaluated medical agent frameworks. The substantial advantage of the agentic reasoning framework over standalone MLLMs can be largely attributed to its ability to invoke specialized perception tools. These tools enable fine-grained extraction of pathological findings and anatomical features, effectively addressing the stringent perceptual requirements of dental image analysis. Additionally, although panoramic radiograph interpretation represents only one component of DentAgent's broader capabilities, DentAgent still achieves better performance than existing medical agents, further demonstrating the effectiveness of our framework for oral and dental reasoning tasks.

\begin{table}[t]
    \centering
    \caption{Comparison on the DentVLM reader-study benchmark. Accuracy and hit rate evaluate multi-class and multi-label diagnosis tasks, respectively.}
    \label{tab:dentvlm_clinical_results}
    \resizebox{.85\linewidth}{!}{%
    \begin{tabular}{lcc}
        \toprule
        \textbf{Model} &
        \textbf{Accuracy (\%) $\uparrow$} &
        \textbf{Hit Rate (\%) $\uparrow$} \\
        \midrule
        
        \multicolumn{3}{@{}l}{\textit{General-purpose MLLMs}} \\
        Qwen3-30A3
        & 50.5
        & 15.1 \\
        Qwen3-235A22
        & 48.7
        & 13.7 \\
        
        \midrule
        \multicolumn{3}{@{}l}{\textit{Medical-specific MLLMs}} \\
        Hulu-30A3
        & 37.6
        & 8.6 \\
        Hulu-235A22
        & 28.7
        & 12.7 \\
        DentVLM
        & \underline{78.8}
        & {55.2} \\
        
        \midrule
        \multicolumn{3}{@{}l}{\textit{Reader Groups}} \\
        Junior Readers
        & 69.8
        & 41.9 \\
        General Practitioners
        & 77.0
        & 55.4 \\
        Senior Specialists
        & \textbf{81.4}
        & \underline{59.3} \\
        
        \midrule
        \textbf{DentAgent (ours)}
        & {72.6}
        & \textbf{76.6} \\
        \bottomrule
    \end{tabular}
    }
\end{table}

\paragraph{DentVLM Reader-study Benchmark}
In the reader-study benchmark of DentVLM, we compared DentAgent with general-purpose MLLMs, medical-specific MLLMs, and reported human readers with varying levels of clinical experience, as shown in Table~\ref{tab:dentvlm_clinical_results}. DentAgent substantially outperforms all baseline models on the multi-label diagnostic task and even exceeds the performance of senior specialists by 17.3 percentage points. Unlike multi-class classification, which typically focuses on assigning a single label for a specific task, multi-label diagnosis requires a comprehensive assessment of the entire image and the identification of as many coexisting oral conditions as possible. Such findings can be easily overlooked even by experienced clinicians. This is precisely where DentAgent benefits from its evidence-driven, iterative multi-round reflection loop, which enables systematic examination and refinement of diagnostic hypotheses. These results also highlight a promising direction for dental AI: rather than replacing clinicians, systems such as DentAgent can serve as adjunctive diagnostic tools by supporting comprehensive image assessment, improving clinical efficiency, and reducing the risk of diagnostic omissions.

\begin{table}[t]
    \centering
    \caption{Comparison on the IOSVQA. For MLLMs without native 3D input support, we project each IOS into multiple 2D views.}
    \label{tab:iosvqa_results}
    \resizebox{.85\linewidth}{!}{
    \begin{tabular}{lcc}
        \toprule
        \textbf{Model} & \textbf{Input Type} & \textbf{Accuracy (\%) $\uparrow$} \\
        \midrule
        
        \multicolumn{3}{@{}l}{\textit{General-purpose MLLMs}} \\
        Qwen3VL-8B
            & 2D multiview
            & 58.84 \\
        InternVL3.5-8B
            & 2D multiview
            & 55.16 \\
        
        \midrule
        \multicolumn{3}{@{}l}{\textit{Medical-specific MLLMs}} \\
        HuluMed-7B
            & 2D multiview
            & 63.56 \\
        HuatuoGPT-V-7B
            & 2D multiview
            & 59.88 \\
        MedGemma-1.5
            & 2D multiview
            & 55.00 \\
        
        \midrule
        \multicolumn{3}{@{}l}{\textit{3D-specific MLLMs}} \\
        PointLLM-7B
            & 3D IOS mesh
            & 46.71 \\
        ShapeLLM-7B
            & 3D IOS mesh
            & 33.18 \\
        IOSVLM
            & 3D IOS mesh
            & \underline{72.42} \\
        
        \midrule
        \textbf{DentAgent (ours)}
            & \textbf{3D IOS mesh}
            & \textbf{78.54} \\
        \bottomrule
    \end{tabular}
    }
\end{table}

\paragraph{IOSVQA}
As shown in Table~\ref{tab:iosvqa_results}, DentAgent outperforms the strongest 3D-specific IOSVLM by 6.12\% and the best-performing 2D multiview baseline by 14.98\%. These substantial gains underscore the value of explicitly grounding reasoning in task-relevant geometric evidence extracted directly from the native 3D mesh. In contrast to 2D multiview approaches, which may lose or distort metric and relational cues during projection, DentAgent operates on the original mesh to quantify inter-tooth spatial relationships, occlusal contacts, and local surface morphology. DentAgent also differs from monolithic 3D models, in which these geometric cues are implicitly compressed into a global representation. Instead, it organizes and aggregates the extracted evidence around the queried tooth or occlusal region. This query-conditioned, target-centric aggregation filters out irrelevant anatomical information while integrating complementary geometric cues, thereby enabling more accurate and reliable answers.

\subsection{Analysis}
\label{sec:analysis}

\begin{figure}[t]
    \centering
    \includegraphics[width=\columnwidth]{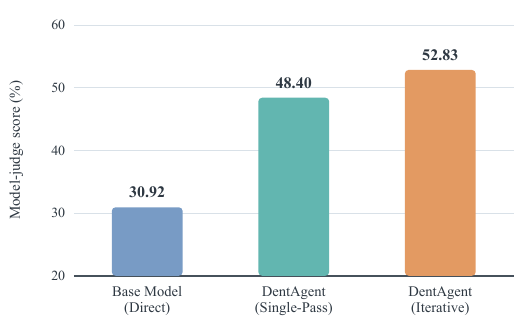}
    \caption{Comparison on MMoralBench between direct inference, single-pass agentic execution, and iterative evidence acquisition under Qwen3.5-9B.}
    \label{fig:framework_ablation}
\end{figure}

\paragraph{Agentic Orchestration Drives the Improvement}
To assess the contributions of different components in DentAgent, we first isolate the effect of agentic orchestration by fixing Qwen3.5-9B as the reasoning backbone across all variants, as shown in Fig.~\ref{fig:framework_ablation}. Compared with direct inference, introducing single-pass agentic execution yields a substantial gain of 17.48\% even before iterative verification is enabled. This result highlights the importance of coordinated multi-agent collaboration and specialized perception and analysis tools for complex multimodal dental reasoning.

\paragraph{Iterative Verification Further Improves Evidence Quality}
With the backbone and toolset unchanged, restoring the complete Identify-Delegate-Observe-Verify loop further raises the score from 48.40 to 52.83. Although single-pass execution can acquire informative observations, it cannot adaptively revisit insufficiently examined targets or resolve evidence gaps and conflicts identified during execution. Taken together, agentic tool use produces the dominant improvement, while reflection loop further enhances the completeness and consistency of the supporting evidence.

\begin{figure}[t]
    \centering
    \includegraphics[width=\columnwidth]{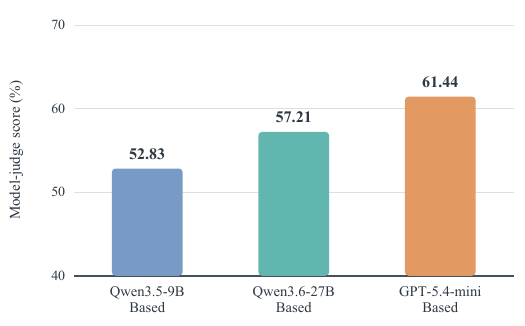}
    \caption{Generalization on MMOralBench of the DentAgent workflow under various reasoning backbones.}
    \label{fig:backbone_comparison}
\end{figure}

\paragraph{Compatibility Across Reasoning Backbones}
We further investigate the effect of various backbone~\cite{qwen3.5,qwen3.6-27b,gpt5} by keeping the DentAgent workflow fixed, as shown in Fig.~\ref{fig:backbone_comparison}. The framework exhibits steady performance gains as the capability of the underlying reasoning backbone increases. This trend suggests that, as a lightweight and training-free dental reasoning workflow, DentAgent can be readily integrated with different backbones while continuing to benefit from advances in their reasoning capabilities. Such compatibility provides practical flexibility for real-world deployment across diverse performance requirements and computational constraints.


\section{Conclusion}
We presented DentAgent, an evidence-centric multi-agent framework that separates evidence acquisition from response generation and coordinates five specialist agents through an Orchestrator and an Evidence Blackboard that preserves source-attributed findings throughout inference. Across four benchmarks, DentAgent demonstrates strong performance across diverse dental modalities and tasks. Future work will extend DentAgent to additional modalities, such as histopathology and CBCT, explore cost-efficient training strategies, and prospectively evaluate its value for clinical support.

\bibliographystyle{IEEEtran}
\bibliography{references}

\end{document}